\documentclass[letterpaper]{article} 
\usepackage{aaai25}  
\usepackage{times}  
\usepackage{helvet}  
\usepackage{courier}  
\usepackage[hyphens]{url}  
\usepackage{graphicx} 
\usepackage{natbib}  
\usepackage{caption} 
\usepackage{algorithm}
\usepackage{algorithmic}
\usepackage{xcolor}

\usepackage{newfloat}
\usepackage{listings}
\DeclareCaptionStyle{ruled}{labelfont=normalfont,labelsep=colon,strut=off} 
\floatstyle{ruled}
\newfloat{listing}{tb}{lst}{}
\floatname{listing}{Listing}
\title{Locally Convex Global Loss Network for Decision-Focused Learning}
\author{
    Haeun Jeon\equalcontrib,
    Hyunglip Bae\equalcontrib,
    Minsu Park,
    Chanyeong Kim,
    Woo Chang Kim\corresponding
}
\affiliations{
    KAIST\\
    \{haeun39, qogudflq, mspark0425, kim.chanyeong, wkim\}@kaist.ac.kr
}

\usepackage{amsmath}
\usepackage{amssymb}
\usepackage{booktabs}
\usepackage{mathtools}
\usepackage{multirow}

\begin{document}
\maketitle
\begin{abstract}
In decision-making problems under uncertainty, predicting unknown parameters is often considered independent of the optimization part.
Decision-focused learning (DFL) is a task-oriented framework that integrates prediction and optimization by adapting the predictive model to give better decisions for the corresponding task.
Here, an inevitable challenge arises when computing the gradients of the optimal decision with respect to the parameters.
Existing research copes with this issue by smoothly reforming surrogate optimization or constructing surrogate loss functions that mimic task loss.
However, they are applied to restricted optimization domains.
In this paper, we propose Locally Convex Global Loss Network (LCGLN), a global surrogate loss model that can be implemented in a general DFL paradigm.
LCGLN learns task loss via a partial input convex neural network which is guaranteed to be convex for chosen inputs while keeping the non-convex global structure for the other inputs.
This enables LCGLN to admit general DFL through only a single surrogate loss without any sense for choosing appropriate parametric forms.
We confirm the effectiveness and flexibility of LCGLN by evaluating our proposed model with three stochastic decision-making problems.
\end{abstract}

\section{Introduction}
\label{sec:intro}

\begin{figure*}[t!]
\begin{center}
\centerline{\includegraphics[width=0.9\textwidth]{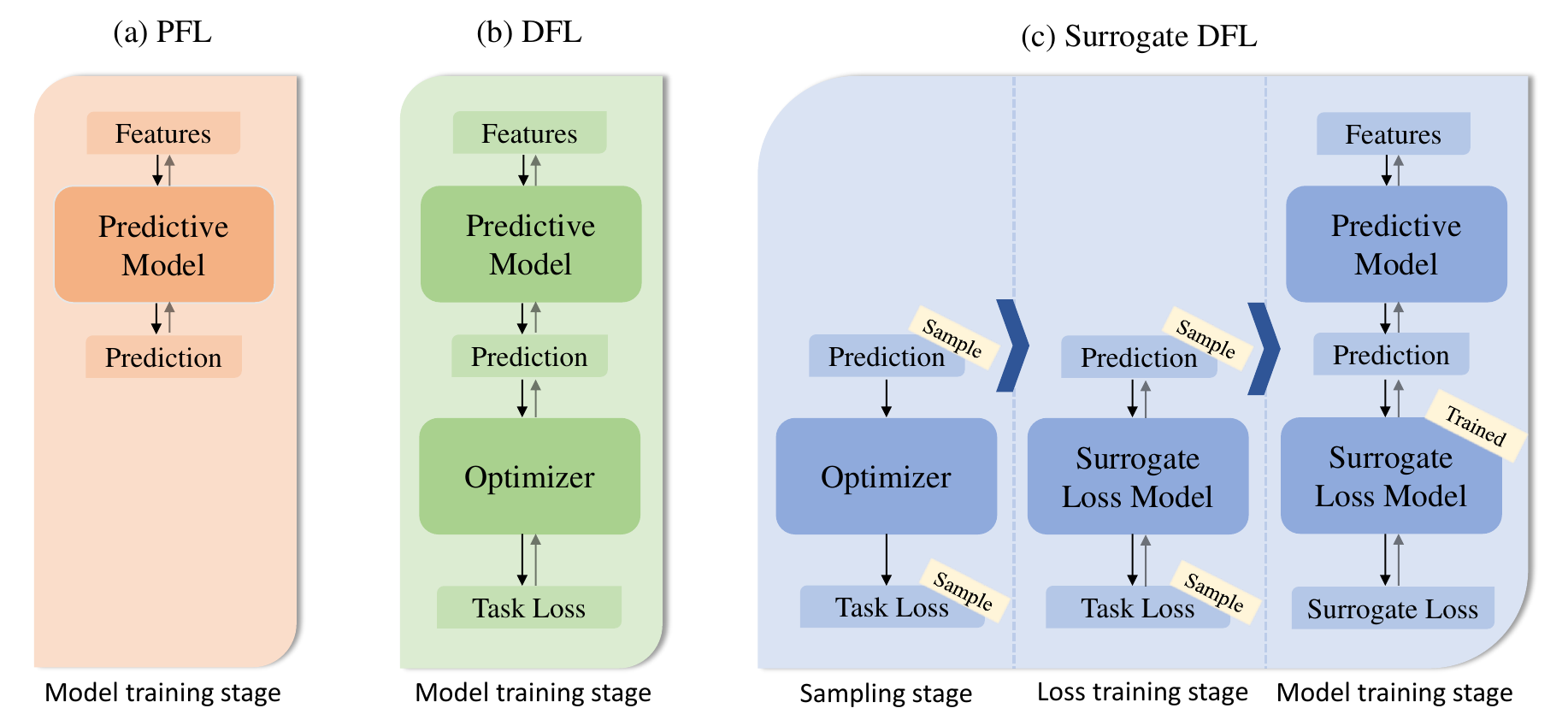}}
\caption{
A model training pipeline for PFL, DFL, and surrogate DFL.
PFL trains the predictive model by minimizing the prediction loss.
DFL directly delivers gradients minimizing the task loss.
Surrogate DFL first learns a surrogate loss model that follows the true task loss by sampling predictions and its task losses.
Then, it trains the predictive model to convey useful gradients derived from the trained surrogate loss model in an end-to-end manner.
}
\label{fig:paradigm}
\end{center}
\end{figure*}

Decision-making problems under uncertainty arise in various real-world applications such as production optimization, energy planning, and asset liability management \cite{shiina2003multistage, carino1994russell, fleten2008short, delage2010distributionally, garlappi2007portfolio}. These problems involve two main tasks: \textit{Prediction} and \textit{Optimization}.
The \textit{Prediction} task aims to create a model for uncertainty and estimate unknown parameters from some input features. This task is often performed using machine learning (ML) techniques such as regression or deep learning.
In the \textit{Optimization} task, the corresponding optimization problem is solved using diverse off-the-shelf solvers with the estimated parameters from the prediction task. For example, in asset liability management, asset returns are estimated in the prediction stage, and the optimal portfolio to repay the liability is obtained in the optimization stage based on these predictions.

Prediction-focused learning (PFL) framework is the most commonly used approach to solve such problems, handling the two tasks in separate and independent steps. PFL first trains the ML model on the input features to produce predictions that closely match the observed parameters. Subsequently, the decision-making problem is defined using the parameters predicted by the trained ML model and is solved to obtain an optimal decision. PFL is based on the underlying belief that accurate predictions will lead to good decisions. However, ML models frequently fail to achieve perfect accuracy, resulting in sub-optimal decisions. Consequently, in many applications, the prediction and optimization tasks are not distinct but are intricately linked and should ideally be considered jointly.

Decision-focused learning (DFL) achieves the above purpose by directly training ML models to make predictions that lead to good decisions.
DFL combines prediction and optimization tasks by creating an end-to-end system that aims to minimize \textit{task loss} directly.
Task loss is defined as the quality of decisions derived from the predictive model and therefore, depends on the optimal solution of the associated optimization problem.
To train an ML model in this context, differentiating through the optimization problem is required to calculate the gradient of the task loss.
This presents a key challenge in integrating prediction and optimization.
However, differentiation may be impossible if the decision variable is discrete or the objective function is discontinuous.

While surrogate optimization still necessitates differentiation of the optimization problem, efforts have been made to develop solver-free surrogate loss functions that can effectively obtain gradients.
We refer to this approach as \textit{Surrogate DFL}.
Recent successful works in Surrogate DFL include SPO \citep{elmachtoub2022smart}, Contrastive Loss \citep{mulamba2020contrastive}, LTR Loss \citep{mandi2022decision}, LODL \citep{shah2022decision}, EGL \citep{shah2024leaving}, LANCER \citep{zharmagambetov2024landscape}, and TaskMet \cite{bansal2024taskmet}.
SPO, Contrastive loss, and LTR loss are specifically designed for linear objectives while LODL can be implemented for general DFL problems.
However, LODL creates and trains a surrogate with a specific parametric form for each data instance resulting in expensive computation, with decision quality heavily dependent on the chosen parametric surrogate form.
While EGL and LANCER have partially addressed the computational cost issue by extending a local model to a global model, the challenge of selecting an appropriate parametric surrogate form still remains.
Alongside the surrogates, TaskMet minimized the prediction error while maintaining the decision quality.
We illustrated the model training pipeline for PFL, DFL, and Surrogate DFL in Figure \ref{fig:paradigm}.

Notably, there have been attempts to construct surrogate loss functions using neural networks \citep{shah2022decision}. However, these approaches generally performed poorly in experiments because naive MLPs fail to capture the local characteristics of the true underlying loss.

In this paper, we propose Locally Convex Global Loss Network (LCGLN), which is a global and general surrogate loss for DFL.
LCGLN adopts partial input convex neural network (PICNN) \cite{amos2017input} as a parametric surrogate form to approximate task loss.
With PICNN, we guarantee the surrogate loss function to be convex for the chosen inputs while maintaining the general structure for the others.
Furthermore, users are not required to possess an artistic sense for selecting the suitable parametric function form for the task loss.
Therefore, LCGLN can handle the general DFL problems using only one surrogate loss, regardless of the number of observed data.
In the experiment section, we demonstrate the capability of LCGLN on three stochastic decision-making problems, namely inventory stock problem, budget allocation problem, and portfolio optimization problem.
We show that LCGLN learns task loss well with a single surrogate loss.

\section{Related Works}
\label{sec:related-works}

Various methodologies based on gradient-based learning in DFL have been developed \cite{mandi2023decision}.
Some approaches directly differentiate the constrained optimization problem to update the model parameters.
For instance, stochastic optimization problems were tackled by directly differentiating convex QPs using KKT optimality conditions and employing Optnet \cite{amos2017optnet} for efficient differentiation \cite{donti2017task}.
For general convex optimization problems, the differentiable solver Cvxpylayers \cite{agrawal2019differentiable} were developed.

Alternatively, some works introduce regularization terms to smooth the optimization mapping.
For example, the Euclidean norm of the decision variables was added to use quadratic programming methods \cite{wilder2019melding}, while the logarithmic barrier term was added to differentiate linear programming problems \cite{mandi2010interior}.
Furthermore, entropy terms were incorporated to solve multi-label problems \cite{martins2017learning, amos2019limited}, while constrained optimization problems were addressed using the perturb-and-MAP \cite{papandreou2011perturb} framework, which added regularization through perturbations \cite{niepert2021implicit, berthet2020learning}.

Recent works on DFL have focused on constructing differentiable surrogate loss models.
Smart "Predict, Then Optimize" (SPO) \cite{elmachtoub2022smart} proposed a convex SPO+ loss where the loss upper bounds the task loss.
To update the predictive model, they obtained a subgradient of their proposed surrogate loss.
NCE \cite{mulamba2020contrastive} used a noise contrastive approach \cite{gutmann2010noise} to build a family of surrogate loss for linear objectives.
LTR \cite{mandi2022decision} applied this approach to ranking problems.
SO-EBM \cite{kong2022end} used an energy-based differentiable layer to model the conditional probability of the decision.
The energy-based layer acts as a surrogate of the optimization problem.
LODL \cite{shah2022decision} used supervised machine learning to locally construct surrogate loss models to represent task loss.
They first sampled predictions near true instances for each instance and built a surrogate loss model respectively.
They proposed three families for the model (e.g. WeightedMSE, Quadratic) to design parametric surrogate loss models.
EGL \citep{shah2024leaving} extends LODL to a global surrogate model and LANCER \citep{zharmagambetov2024landscape} learns such handcraft global surrogate loss via actor-critic framework.

\section{Preliminaries}
\label{sec:pre}

In this section, we motivate the necessity of DFL approach from a simple knapsack example.
We also briefly summarize surrogate loss models and input convex neural networks.

\begin{figure}[t!]
\centering
\includegraphics[width=\columnwidth]{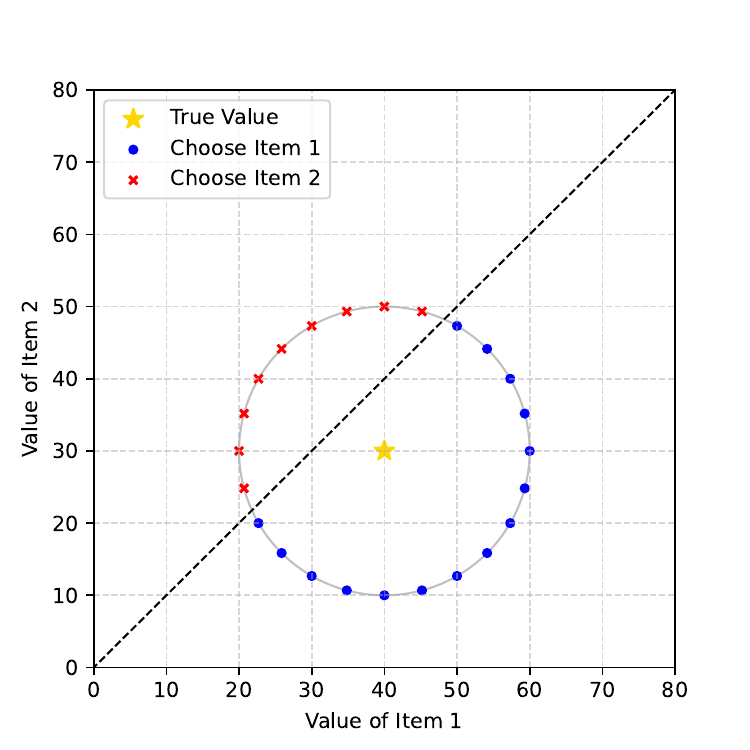}
\caption{
A simple example of a knapsack problem.
There are two items valued \$40, \$30 each, marked with a yellow star.
We predict the value of items and choose the higher one.
Blue dots and red crosses are predicted values representing good and bad decisions respectively.
PFL gives the same prediction loss for every prediction while DFL gives \$10 loss in red cross and 0 in blue. 
}
\label{fig:knapsack}
\end{figure}

\paragraph{Comparison on PFL and DFL}
Prediction-focused learning (PFL) and decision-focused learning (DFL) are two big learning pipelines for decision-making problems under uncertainty as in Figure \ref{fig:paradigm}.
While PFL learns the predictive model focusing on the prediction from the predictive model, DFL focuses on the objective of the downstream optimization problem, commonly referred to \textit{task loss}.
Starting with PFL, the model training step is divided into two stages, prediction and optimization.
In the prediction stage, PFL learns a predictive model by minimizing the prediction loss such as MSE.
In the optimization stage, it solves an optimization problem using the prediction of the predictive model as parameters.
By contrast, the predictive model in DFL is trained in an end-to-end manner.
It prioritizes learning the model to make good decisions (or actions) by minimizing the task loss, rather than optimizing the prediction loss.

We introduce a motivating example with a simple knapsack problem in Figure \ref{fig:knapsack} to give motivation in using DFL. 
Let's consider two items: item 1 with a value of \$40 and item 2 with \$30.
Two axes in Figure \ref{fig:knapsack} represent the value of each item.
The true optimal value $(40,30)$ is marked by a yellow star.
Assume we can only select one item.
Our objective is to predict the value of items by only observing the item features and choosing one with a higher value.
If we choose item 1 (the optimal decision), we would be satisfied, earning \$40.
Conversely, choosing item 2 would be a sub-optimal decision, resulting in a relative loss of \$10.
Suppose we predict the values as blue dots and red crosses.
From a PFL point of view, the prediction loss is equal to \$20.
Any prediction on the gray circle results in the same prediction loss, which is not informative as the blue dots are correct decisions, whereas the red crosses are not. 
From a DFL standpoint, blue dots return zero task loss while red crosses give task loss of \$10.
This example shows that minimizing the prediction loss \emph{may not} lead to lower task loss. 

\paragraph{Surrogate DFL}
DFL computes task loss gradients with respect to the input parameters by, for example, directly differentiating the optimization problem (or optimizer) \cite{donti2017task}.
However, the optimizer cannot be differentiated easily in most cases.
To tackle this, methods using surrogate loss functions were proposed \cite{elmachtoub2022smart}.
Surrogate loss models are differentiable and it approximates the mapping between the prediction and the task loss.
Consequently, it can be used to calculate the gradient to update the predictive model efficiently.

Given dataset $\mathcal{D}=\{ (x_{i}, y_{i}) \}^{N}_{i=1}$ with $N$ instances, our goal is to minimize the regret $\mathcal{R}$ for the optimization problem:
\begin{equation}
\label{eq:regret}
    \begin{aligned}
     & \min_{\theta} & \quad \mathcal{R}(\hat{y},y) & \vcentcolon= \mathcal{L}_{task}(a^*(\hat{y}) , y) - \mathcal{L}_{task}(a^*(y) , y) \\
     & \textrm{s.t.} & \quad \hat{y} & = \mathcal{M}_{\theta}(x) \\
    & & a^{*}(\hat{y}) & = \operatorname*{arg\,min}_{a \in \mathcal{A}} \mathcal{L}_{task} (a , \hat{y})
    \end{aligned}
\end{equation}
where $\mathcal{L}_{task}$ is the task loss to be optimized, $\hat{y}$ is the prediction from the predictive model $\mathcal{M}$ parameterized by $\theta$ and $a^* \in \mathcal{A}$ is an optimal action (or decision) in a feasible region $\mathcal{A}$ derived by any off-the-shelf solver.
Note that the second term of the objective is nothing but a constant optimal loss, and therefore our objective is equivalent to minimizing $\mathcal{L}_{task} (a^*(\hat{y}), y)$.

To update the predictive model via gradient descent w.r.t. $\theta$, one must find the gradient 
$ \partial \mathcal{R} / \partial \theta $.
Using the chain rule, the gradient can be decomposed into:
\begin{align}
    \frac{ \partial \mathcal{R}( \hat{y}, y) }{ \partial \theta }
    &=
    \frac{ \partial \mathcal{R}( \hat{y}, y) }{ \partial \hat{y} }
    \cdot
    \frac{ \partial \hat{y} }{ \partial \theta } \\
    &\approx
    \frac{ \partial \mathcal{L}_{\psi}(\hat{y}, y) }{ \partial \hat{y} }
    \cdot
    \frac{ \partial \hat{y} }{ \partial \theta }
\end{align}
where $\mathcal{L}_{\psi}(\hat{y}, y)$ is the parametric surrogate loss with parameter $\psi$ that is trained to approximate $\mathcal{R}(\hat{y}, y)$.
The global surrogate loss can be trained to richly approximate the true task loss as:

\begin{equation}
    \psi^* = \operatorname*{arg\,min}_{\psi}\mathop{\mathbb{E}}_{\hat{y},y} [ \lVert \mathcal{L}_{\psi}(\hat{y},y) - \mathcal{R}( \hat{y}, y) \rVert ]
\end{equation}
Training the predictive model $\mathcal{M}_{\theta}$ can be easily done with backpropagation using the gradient $ \partial \mathcal{L}_{\psi} / \partial \theta $.

\paragraph{Partial Input Convex Neural Network}
Amos et al. \cite{amos2017input} proposed an input convex neural network (PICNN) that ensures convexity with respect to the chosen inputs.
PICNN is constructed by introducing additional weights to connect the input layer to each hidden layer. 
Given this structure, non-decreasing convex activation functions such as softplus and non-negativity of weights connecting between hidden layers are required to guarantee the convexity of PICNN.
LCGLN uses PICNN to richly represent the non-convex nature of the true loss mapping, and simultaneously ensure local convexity.

\section{Locally Convex Global Loss Network}
\label{sec:model}
In various optimization problems, each has its own specific objectives known as \emph{task losses}.
Our goal is to devise a surrogate loss function capable of accurately representing the true task loss across a range of optimization problems. In this paper, we introduce LCGLN, where a single loss representation can replace the true task loss. While DFL methods suffer from differentiating through the optimization solver when the optimization problem is not smooth, our LCGLN is easily differentiable and therefore can be readily backpropagated when updating the predictive model via gradients.


\begin{algorithm}[t]
   \caption{Training Predictive Model with LCGLN $\mathcal{L}_{\psi}$}
   \label{alg:LCGLN}
\renewcommand{\arraystretch}{1.2}
\begin{algorithmic} \small
    \STATE {\bfseries Dataset:}  $\mathcal{D}=\{ (x_{i}, y_{i}) \}^{N}_{i=1}$
    \STATE {\bfseries 1. Generate Samples:}
    \STATE Sample set $\mathcal{S} \leftarrow \emptyset$
    \FOR{$i=1,\ldots,N$}
        \STATE $\mathcal{S} \leftarrow \mathcal{S} \cup (y_i, y_i, 0)$
    \ENDFOR
    \STATE Initialize sampling model $\mathcal{M}_{\xi}$.
    \FOR{ $k=1,\ldots,K-1$ }
        \FOR{$i=1,\ldots,N$}
            \STATE $ \tilde{y}_i^{(k)} = \mathcal{M}_{\xi}(x_i) $
            \STATE $\xi \leftarrow$ Update$(\xi, \nabla_{\xi} \lVert \tilde{y}_i^{(k)} - y_i \rVert )$
            \STATE Solve $a^{*(k)}_i = \operatorname*{argmin}_{a} \mathcal{L}_{task} (a(\tilde{y}_i^{(k)}), y_i)$.
            \STATE $\mathcal{S} \leftarrow \mathcal{S} \cup (\tilde{y}_i^{(k)}, y_i, \mathcal{R}(\tilde{y}_i^{(k)}, y_i) )$
        \ENDFOR
    \ENDFOR

    \STATE {\bfseries 2. Learn LCGLN $\mathcal{L}_{\psi}$:} 
    \STATE Initialize surrogate loss model $\mathcal{L}_{\psi}$.
    \FOR{$(\hat{y}, y, \mathcal{R}(\hat{y},y))$ in $\mathcal{S}$}
    \STATE $\psi \leftarrow$ Update $(\psi, \nabla_{\psi} \lVert \mathcal{L}_{\psi}(\hat{y},y) - \mathcal{R}( \hat{y}, y) \rVert )$
    \ENDFOR
    
    \STATE {\bfseries 3. Train Predictive Model $\mathcal{M}_{\theta}$:}
    \STATE Initialize predictive model $M_{\theta}$.
    \STATE $\theta \leftarrow$ Update $(\theta, \nabla_{\theta}\hat{y} \cdot \nabla_{\hat{y}} \mathcal{L}_{\psi}(\hat{y},y))$

\end{algorithmic}
\end{algorithm}

We now introduce the end-to-end training procedure for LCGLN.
The goal is to learn a regret mapping $(\hat{y}, y) \rightarrow \mathcal{R}(\hat{y},y)$ to obtain gradient $ \partial \mathcal{R} / \partial \theta $ for updating the predictive model $\mathcal{M}_{\theta}$.
The training consists of three main steps: generating samples, learning a global surrogate loss, and training a predictive model.

\paragraph{Generating Samples}
To train the LCGLN in a supervised learning manner, we generate $K$ samples for each $N$ instance, i.e. for each instance $y_i$, we sample $ \tilde{y}_i^{(1)}, \tilde{y}_i^{(2)}, ..., \tilde{y}_i^{(K)} $.

Some previous research assumed that predictions $\tilde{y}$ would closely approximate the true labels $y$, typically generating sample predictions by simply adding Gaussian noise.
However, the Gaussian sampling method may be challenging to apply for two reasons.
First, in some cases, the true label $y$ is unknown.
For example, in the inventory stock problem used in our experiments, the true $y$ represents the probability vector of each demand occurring.
Yet, the observed data only provides a specific demand value realized according to that probability.
In such cases, since the true $y$ is unknown, there is no target to which Gaussian noise can be added.
Second, even if the true $y$ is known, determining the appropriate standard deviation for sampling around $y$ can be difficult, often requiring repeated tuning.

On account of this, we adopt the model-based sampling (MBS) approach \cite{shah2024leaving}.
MBS involves constructing a sampling model $M_{\xi}$ that mirrors the architecture of the predictive model and training within a PFL paradigm using MSE.
During the training, we take inferences from the intermediate model and generate samples.
Unlike the Gaussian sampling, neither the true \( y \) nor the standard deviation for the noise is required for MBS.

\begin{table*}[t]
    \centering
    \begin{small}
    \begin{sc}
    \renewcommand{\arraystretch}{1.3}
        \begin{tabular}{clccc}
        \toprule
        \multirow{2}{*}{\textbf{Paradigm}} & \multirow{2}{*}{\textbf{Methods}} & \multicolumn{3}{c}{\textbf{Problem}} \\
        \cline{3-5}
               &  & Inventory  & Budget (500 Fakes)  & Portfolio  \\
        \midrule
        2 Stage & \texttt{PFL}      & 0.242 $\pm$ 0.005 & 0.513 $\pm$ 0.016 & 0.189 $\pm$ 0.002  \\
        \midrule
        Exact Diff & \texttt{DFL}      & 0.228 $\pm$ 0.002 & 0.532 $\pm$ 0.020 & 0.187 $\pm$ 0.002  \\
        \midrule
        Local Surrogate & \texttt{LODL-DQ}  & 0.378 $\pm$ 0.007 & 0.503 $\pm$ 0.020 & 0.193 $\pm$ 0.002  \\
        \midrule
        \multirow{4}{*}{Global Surrogate} & \texttt{LANCER}   & 0.182 $\pm$ 0.004 & 0.490 $\pm$ 0.010 & 0.246 $\pm$ 0.008  \\
         & \texttt{EGL-WMSE} & 0.371 $\pm$ 0.002 & 0.510 $\pm$ 0.013 & 0.187 $\pm$ 0.001  \\
         & \texttt{EGL-DQ}   & 0.369 $\pm$ 0.007 & 0.492 $\pm$ 0.005 & 0.256 $\pm$ 0.002  \\
         & \texttt{LCGLN}     & \textbf{0.174 $\pm$ 0.002} & \textbf{0.468 $\pm$ 0.009} & \textbf{0.185 $\pm$ 0.000}  \\
        \bottomrule
        \end{tabular}
    \end{sc}
    \end{small}
    \caption{
    The table contains normalized test regret $\mathcal{R}_{test} / \mathcal{R}_{worst}$ with standard error mean (SEM) tested on three stochastic optimization problems.
    The metric is \emph{lower the better} with an optimal regret of zero.
    The best-performing results for each problem are bold-lettered.
    We evaluate PFL, DFL, local, and global surrogate loss models.
    We use 32 samples for the surrogate loss models. 
    The global surrogate loss LCGLN outperformed the baselines across all three problems.
    }
    \label{table:main-table}
\end{table*}

\paragraph{Learning Global Surrogate Loss LCGLN}

Our objective is to learn a mapping $(\hat{y},y) \rightarrow \mathcal{R}$ for conveying informative gradients of task loss.
We propose leverage of partial input convex neural network (PICNN) \cite{amos2017input} as our Locally Convex Global Loss Network (LCGLN).
Our motivation for using PICNN as a surrogate model for task loss is fourfold:
\begin{itemize}
    \item \textit{Expressiveness: }
        A good surrogate loss model should have a sufficient number of parameters to accurately approximate the true task loss since a lack of expressiveness may lead to under-performance.
        \citet{amos2017input} proved that a $k$-layer PICNN can represent any $k$-layer feedforward network. Since feedforward neural networks are known as universal approximators \cite{cybenko1989approximation, funahashi1989approximate, hornik1991approximation}, this ensures the PICNN can accurately model task loss.
    \item \textit{Easily Differentiable: }
        We need to differentiate our loss model with respect to $\hat{y}$ for gradients training the predictive model.
        Discontinuous or non-differentiable loss models face significant challenges in such tasks.
        In contrast, PICNN can be easily differentiated by using built-in backward functions.
    \item \textit{Locally Convex, but not Globally: }
        We want the model to capture the highly non-convex structure of the true underlying task loss mapping.
        Task loss can be expressed as $f(x^\ast(\hat{y}), y)-f(x^\ast(y), y)$, where it achieves its minimum regret zero when $\hat{y}=y$.
        While task loss may not be convex in $\hat{y}$ for a given $y$ in general, we use PICNN to drive $\hat{y}$ towards the true value $y$ by eliminating local minima except $y$.
        This property enables the model to provide informative gradients for training the predictive model with a small sample size.
        Furthermore, we induce the local minima by adding $\{(y_i, y_i, 0)\}_{i=1}^{N}$ to the dataset for every instance.
    \item \textit{Generality: }
        The adoption of PICNN helps generalize the function approximators and allows the approximation of differentiable optimization.
        Using PICNN, we can only focus on the overall network architecture and perform a simple hyperparameter search instead of requiring well-trained experts' efforts in choosing the right parametric function forms for specific problems.
        
\end{itemize}

\paragraph{Training Predictive Model}
The predictive model $M_{\theta}$ learns a mapping $x \rightarrow y$.
To obtain gradients for the predictive model training, we first generated samples using the MBS approach.
For each generated sample, we derived the true task loss in regret form and added the sample, instance, and regret pair to the dataset.
Additionally, we included $\{(y_i, y_i, 0)\}_{i=1}^{N}$ into the dataset to induce local minima for every instance $y_i$.
Using this dataset, we trained the global surrogate loss model structured with PICNN.

Building on these steps, we now train the predictive model $M_{\theta}$ via gradient descent utilizing the gradients provided by the global surrogate loss $\mathcal{L}_{\psi}$.
The entire procedure from generating the dataset to training the predictive model is summarized in Algorithm \ref{alg:LCGLN}.

\section{Experiments and Results}
\label{sec:exp}

We validate our methodology with three stochastic optimization problems.

\subsection{Experimental Settings}
\label{subsec:exp-settings}
In this section, we explain the experimental details and the baselines used to evaluate our method.
Each problem is elaborated in two stages: the parameter prediction stage and the optimization stage.
\begin{figure*}[t]
\centering
\includegraphics[width=\textwidth]{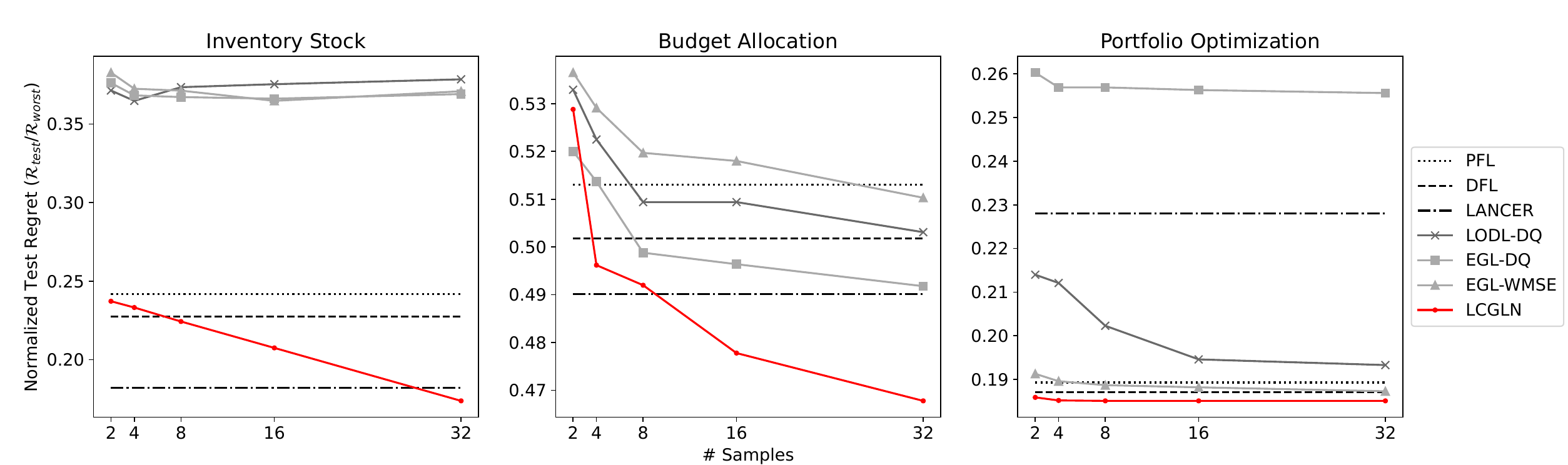}
\caption{
A line plot showing the normalized test regret $\mathcal{R}_{test} / \mathcal{R}_{worst}$ for each methodology and problem setting.
The metric is \emph{lower the better}, with 0 representing the optimal value. 
We tested sample size of $\{2,4,8,16,32\}$ for each problem.
The standard error mean for each experiment is detailed in Appendix \ref{subsec:appen-diff-smpsize}.
Our global surrogate loss LCGLN represented by the red straight line outperforms when 32 samples are used.
}
\label{fig:diff-samplesize}
\end{figure*}
\paragraph{Problem Description}
We conducted experiments on three different stochastic optimization problems, each presenting unique challenges as explored in previous research \cite{donti2017task, wilder2019melding, shah2022decision}.
Our experiments were built on top of the public codes from previous research \cite{donti2017task, shah2022decision}.
Here we briefly describe each problem.
For further problem descriptions, please refer to Appendix \ref{subsec:appen-exp}.

\begin{itemize}
    \item \textbf{Inventory Stock} \cite{donti2017task}:
    We decide the order quantity to minimize the cost over the stochastic demand.
    To simplify the problem, we assume the demands are discrete.
    
    \textit{Prediction}:
    We predict the discrete probability distribution of the demand. 
    
    \textit{Optimization}:
    With the predicted demand distribution, we decide the order quantity that minimizes the cost.

    \item \textbf{Budget Allocation} \cite{wilder2019melding}:
    We choose websites to advertise based on click-through rates (CTRs) of users among websites.
    As a variant, we conduct four different experiment settings by concatenating $\{0, 5, 50, 500\}$ size of random CTRs (fake targets) to the original CTRs and adjusting the problem difficulty.
    
    \textit{Prediction}:
    Given the website features, we predict the CTRs.
    
    \textit{Optimization}:
    With the predicted CTRs, we choose the websites to advertise that maximize the expected number of users who click on the advertisement at least once.
    
    \item \textbf{Portfolio Optimization} \cite{shah2022decision}:
    We allocate weights for invested stocks to maximize the expected risk-adjusted return.
    
    \textit{Prediction}:
    Given the historical daily stock return data, we predict the future stock price.
    
    \textit{Optimization}:
    Using the predicted stock price, we assign portfolio weights that maximize the expected risk-adjusted return, given the covariance matrix.
    
\end{itemize}

\begin{figure*}[t]
\centering
\includegraphics[width=\textwidth]{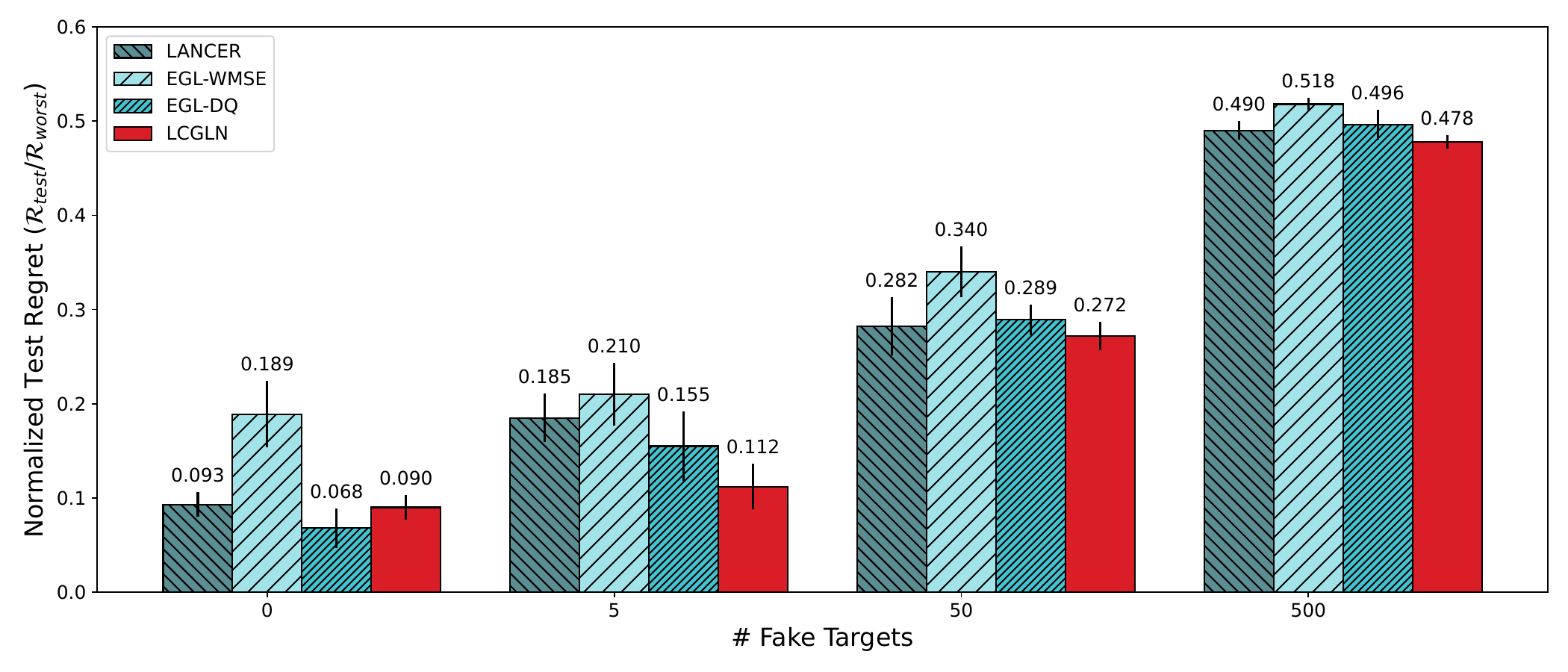}
\caption{
A histogram presenting normalized test regret $\mathcal{R}_{test} / \mathcal{R}_{worst}$ with standard error mean (SEM) for global surrogate loss models in budget allocation with varying number of fake targets.
We use 16 samples for learning loss.
The metric is \emph{lower the better} and 0 when optimal.
We test with $\{0,5,50,500\}$ fake targets, noting that the problem becomes more challenging as the number of fake targets increases.
Our LCGLN shown in red bars outperforms most settings.
}
\label{fig:budget-diff-fake}
\end{figure*}

\paragraph{Baselines}
We compare our global surrogate loss network LCGLN with the following baselines from previous research.
When available, we utilized the hyperparameters specified in previous research for the baselines.
In cases where these were not provided, we endeavored to fine-tune them to perform their best.

\begin{itemize}
    \item \textbf{PFL}:
        A standard approach that trains a predictive model on the input features to produce predictions that closely match the observed parameters.
        We use the negative log-likelihood (NLL) loss for the inventory stock problem and the mean squared error (MSE) loss for the other problems.
        
    \item \textbf{DFL}:
        A direct approach that differentiates through the solver to derive gradients for training the predictive model.
        For the inventory stock and the portfolio optimization problems, we use the differentiable QP solver \cite{donti2017task}.
        We use multilinear relaxation \cite{wilder2019melding} for the budget allocation problem.
             
    \item \textbf{LODL}:
        The local surrogate model \cite{shah2022decision} that learns a surrogate loss locally for each instance, creating total $N$ surrogate loss models.
        To learn a loss model, LODL generates $K$ number of samples for each $N$ instance and solves for the input of the model.
        We use directed-quadratic surrogate loss (LODL-DQ) from their paper, the most promising result among the proposed 4 different loss models.


    \item \textbf{LANCER}:
        The global surrogate model \cite{zharmagambetov2024landscape} that uses alternating optimization to learn the predictive model and the global surrogate loss model.
        The prediction samples are sampled from the predictive model and are used to train the surrogate loss.
        The predictive model is then trained using the gradients from the updated surrogate loss.
        
    \item \textbf{EGL}:
        The global surrogate model \cite{shah2024leaving} that learns a global surrogate loss with feature-based parameterization and model-based sampling.
        We use directed-quadratic (EGL-DQ) and weighted-MSE (EGL-WMSE) for the baseline, which showed the best results on the budget allocation and the portfolio optimization problems respectively. 
        
\end{itemize}

\paragraph{Evaluation Metric}
We use the normalized test regret $\mathcal{R}_{test} / \mathcal{R}_{worst}$ as the evaluation metric, where $\mathcal{R}_{test}$ is a test regret with regret defined in Equation \ref{eq:regret}.
For the maximization tasks, such as the budget allocation and portfolio optimization problem, regrets are calculated by multiplying the negative sign on the corresponding objective. 
We calculate the worst case regret $\mathcal{R}_{worst}$ for each problem and derive the normalized test regret.
For the inventory stock problem, we assumed the worst-case scenario when the company ordered no stock.
In the budget allocation problem, the worst-case scenario was assumed when the advertisements were allocated to the lowest predicted CTRs.
For the portfolio optimization problem, we considered the worst-case scenario to be when the entire investment was allocated to the stock with the lowest predicted return.
Our metric is 0 when optimal and 1 when worst.

We used a predictive model $\mathcal{M}_{\theta}$ as one hidden layer MLP and a learning rate of 0.001.
500 hidden nodes were used for the portfolio optimization and 10 for the other problems.
We mirrored the predictive model exactly for the sampling model $\mathcal{M}_{\xi}$.
For LCGLN, we employed a single hidden layer PICNN with two nodes per layer, a learning rate of 0.001, and a softplus activation function.
A figure illustrating the LCGLN can be found in Appendix \ref{subsec:appen-LCGLN-fig}.
For each global surrogate loss model employing model-based sampling, we selected the learning rate that demonstrated the best performance from $\{0.01,0.05,0.1,0.5,1\}$.
We run 10 experiments for each setting to ensure statistical significance.
The experiments were performed on a Ryzen 7 5800X CPU and an RTX 3060 GPU with 64GB of RAM.

\subsection{Results}
\label{subsec:results}

Table \ref{table:main-table} shows the normalized test regret $\mathcal{R}_{test} / \mathcal{R}_{worst}$ with standard error mean (SEM) for the inventory stock, budget allocation with 500 fake targets and portfolio optimization problems.
For the surrogate loss models, we used 32 samples to learn the loss.
We categorize the methods into four major training paradigms: two-stage, exact differentiation, local surrogate loss, and global surrogate loss.

In the inventory stock problem, LODL-DQ, EGL-WMSE, and EGL-DQ suffer from high regret, indicating that they do not provide informative gradients.
LANCER and LCGLN perform well compared to others due to their \emph{expressiveness}, as both use neural networks to learn the loss.
For the budget allocation problem, we conduct four experiments with $\{0,5,50,500\}$ fake targets.
Detailed results with different fake targets are available in Appendix \ref{subsec:appen-diff-faketargets}.
For the hardest problem with 500 fake targets, the surrogate models tend to show better results than PFL or DFL methods.
In the portfolio optimization problem, good prediction in parameters showed better decisions.
This is evidenced by PFL and DFL showing better results compared to surrogate models.
EGL-WMSE and LCGLN also demonstrated strong performance in our settings.

To reduce the computational cost, it is crucial to train the surrogate loss with small sample sizes.
We conducted experiments with varying sample sizes, as shown in the line plot in Figure \ref{fig:diff-samplesize}.
Note that the normalized test regret is better when lower.
Since PFL, DFL, and LANCER do not vary with changes in a number of sample predictions, they show consistent results across all sample sizes.
The surrogate models showed decreased regret as the sample size increased.
At a sample size of 32, LCGLN outperformed the provided baselines.
For detailed experimental results for each sample size and problem, please refer to Appendix \ref{subsec:appen-diff-smpsize}.

We also tested different numbers of fake targets, varying in $\{0,5,50,500\}$, to show how models perform on harder problems.
We used 16 samples to train the loss model.
Figure \ref{fig:budget-diff-fake} presents the normalized test regret with SEM for global models across each experimental setting.
LCGLN consistently outperformed in most settings.

\section{Conclusion}
\label{sec:conclusion}

In this paper, we propose Locally Convex Global Loss Network (LCGLN), a global and general surrogate loss for DFL. 
LCGLN utilizes PICNN as a parametric surrogate to approximate task loss.
We use PICNN to guarantee the surrogate to be convex near instances while maintaining a general non-convex structure globally.
With LCGLN, users do not need to manually select the appropriate parametric function form for the task loss. Consequently, LCGLN can address general DFL problems with a single surrogate loss, regardless of the amount of observed data.
We evaluated our method on three stochastic optimization problems, achieving better decision quality with fewer training samples compared to the state-of-the-art baselines.
However, despite the expressive power of LCGLN, a limitation remains: achieving high decision quality requires careful selection of the samples.
Thus, our future research will focus on identifying sampling strategies to improve the decision quality for surrogate loss models.

\section*{Acknowledgements}
\label{sec:sasa}
This research was supported by the National Research Foundation of Korea (NRF) grant funded by the Korean government (MSIT) (NRF-2022M3J6A1063021 and NRF-RS-2023-00208980).

\bibliography{ref}

\appendix
\onecolumn

\section{Locally Convex Global Loss Network Details}

\subsection{Architecture Details}
\label{subsec:appen-LCGLN-fig}

Here we show the architecture details of our Locally Convex Global Loss Network (LCGLN).
The input is given as the pair of the sampled prediction $\hat{y}$ and its corresponding instance $y$.

The bluish boxes represent the vectors that ensure convexity with respect to the input, while the reddish boxes correspond to the vectors that do not guarantee convexity. 
The yellow boxes denote the hidden layers, where the superscript inside the parentheses indicates the dimension, and the subscript identifies the layer number.
For instance, $h_i^{(z)}$ is an $i$-th hidden layer with $h_i^{(z)} \in \mathbb{R}^z$.

\begin{figure*}[ht]
\begin{center}
\centerline{\includegraphics[width=\textwidth]{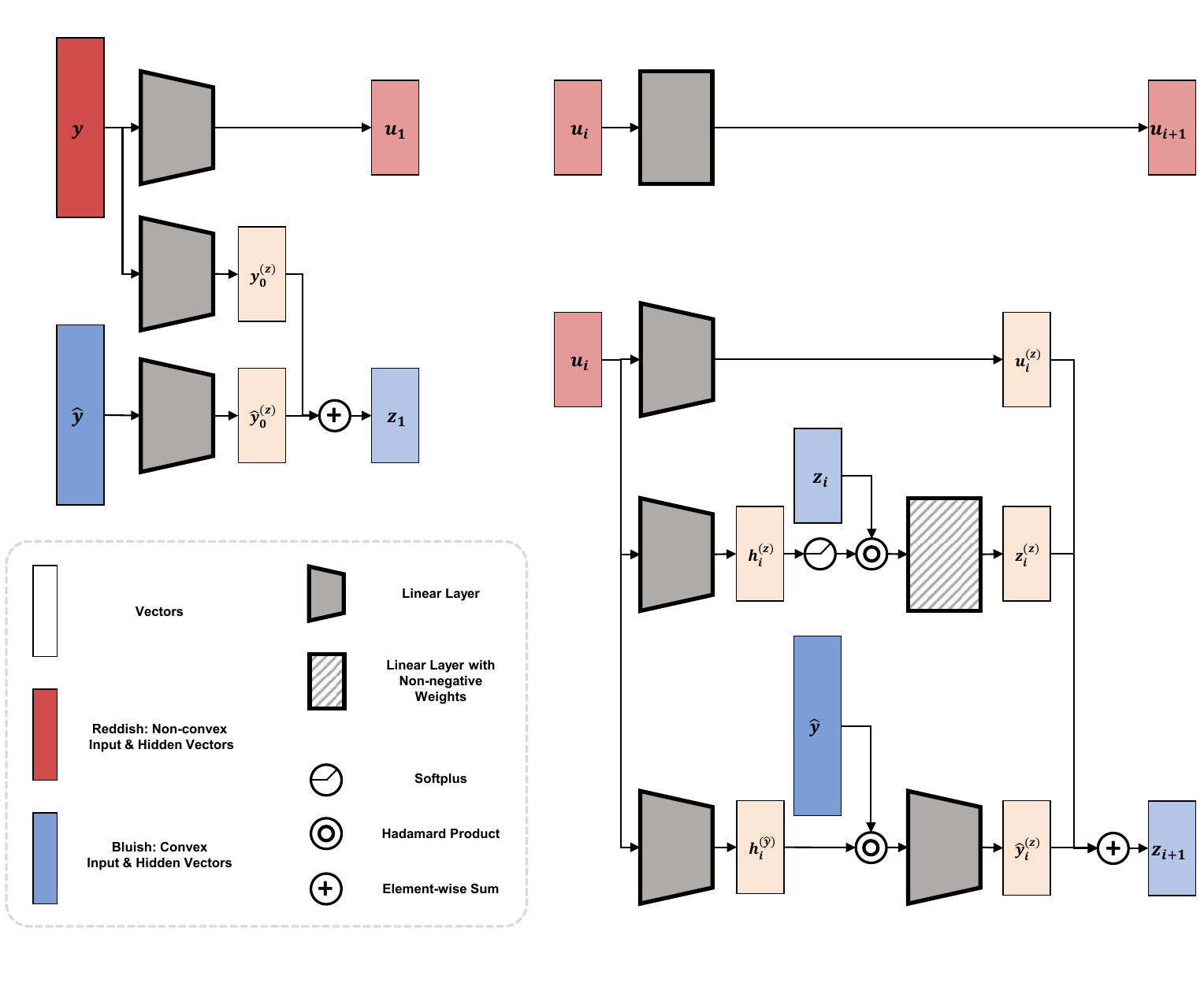}}
\caption{
Architecture of Locally Convex Global Loss Network.
The left figure represents the input to the first hidden layer.
The right figure shows the hidden to hidden layers for non-convex vectors (reddish, upstream) and convex vectors (bluish, downstream).
The output layer can simply be considered as $z_{i+1}$ without $u_{i+1}$.
}
\label{LCGLN-architecture}
\end{center}
\end{figure*}

We depict linear layers as trapezoids in gray when the dimensions change and as rectangles when they remain constant.
The hatched box represents the linear layer with non-negative weights.
Details for operators are available in Figure \ref{LCGLN-architecture}.

The left side of the figure shows the architecture from the input to the first hidden layer.
The right side presents how hidden layers are composed.
The upstream of the right side is a flow for reddish vector representations, while the downside is for bluish vectors.
For the output layer, one may simply let $z_{i+1}$ be the output vector and ignore $u_{i+1}$.

\section{Experimental Details}
\label{subsec:appen-exp}

\subsection{Codes for Paper}
The codes for experiments in our paper are available at the following link.
\begin{links}
  \link{Code}{https://github.com/HESEDaniel/lcgln}
\end{links}

\subsection{Details in Inventory Stock Problem}

In the inventory stock problem \cite{donti2017task}, we consider a situation where one must order $a$ number of products before the demand $y$ is realized to minimize the cost.
We generate problem instances from randomly sampled $x\in \mathbb{R}^{20}$ with the non-linear mapping $p(y\vert x;\theta)\propto \text{exp}(\Theta^\top x)$ for the fixed true parameter $\Theta \in \mathbb{R}^{20 \times 5}$.
To simplify the problem, we assume the demands are discrete with values $d_1, d_2, ..., d_k$ with probabilities $(p_{\theta})_i \equiv p(y=d_i \vert x;\theta) $.
Note that the realization $y$ of $p_{\theta}$ is known while the true distribution $p_{\theta}$ is unknown.
Since, the true distribution $p_{\theta}$ is unknown, we transform $y$ to a $p_{\theta}$ form when training the surrogate loss functions.

The cost function is defined as:
\begin{equation}
    \begin{aligned}
     f(y,a) = c_0a+\frac{1}{2}q_0 a^2 + c_b[y-a]_+ + \frac{1}{2}q_b([y-a]_+)^2+c_h[a-y]_+ + \frac{1}{2}q_h([a-y]_+)^2
    \end{aligned}
\end{equation}
where our objective is to minimize the expected cost:
\begin{equation}
    \begin{aligned}
     \min_{a} \quad \boldsymbol{E}_{y\sim p(y\vert x;\theta)}\left[f(y,a)\right] .
    \end{aligned}
\end{equation}

Here, we let $k=5$ and assign $d_1, \ldots, d_5$ as 1, 2, 5, 10, 20 with probability $(p_{\theta})_i \equiv p(y=d_i\vert x;\theta)$ for each discrete demand; identical to the previous research settings \cite{donti2017task}.
Then, we can rewrite the optimization problem as:
\begin{equation}
    \begin{aligned}
     \min_{a\in\mathbb{R},\quad a_b,a_h\in\mathbb{R}^k} \quad\quad & c_0a +\frac{1}{2}q_0a^2 + \sum_{i=1}^k (p_{\theta})_i\left( c_b(a_b)_i+\frac{1}{2}q_b(a_b)_i^2+c_h(a_h)_i+\frac{1}{2}q_h(a_h)_i^2\right)\\
     \text{s.t.} \quad\quad\quad\quad & d-a\boldsymbol{1}\leq a_b,\;\, a\boldsymbol{1}-d\leq a_h,\;\, a,a_h,a_b\geq 0.
    \end{aligned}
\end{equation}

The parameters $c$ and $q$ each represent the coefficient of linear and quadratic costs, respectively.
Specifically, $c_0$ and $q_0$ are related to the costs of the amount of product ordered.
$c_b, q_b$ and $c_h, q_h$ are related to the amount of under and over-ordered. 
We use the setting $(c_0, q_0, c_b, q_b, c_h, q_h)$ with $(10, 2, 30, 14, 10, 2)$.


\subsection{Details in Budget Allocation Problem}

The objective of the submodular optimization problem \cite{wilder2019melding} is to choose $B=2$ websites to advertise based on click-through rates (CTRs) of $U=10$ users among $W=5$ websites.
We generate the website features $x_w \in \mathbb{R}^{U}$ for each website $w$ by multiplying a random matrix $A \in \mathbb{R}^{U \times U}$ to the CTRs $y_w \in \mathbb{R}^{U}$.
We conduct experiments with $\{0, 5, 50, 500\}$ fake targets to add difficulty by concatenating the fake target size of random CTRs to the original CTRs.
Given the website features $x_w$ for each website $w$, we first predict the CTRs $\hat{y}_w$.
Then, we choose $B$ websites to advertise using a binary vector $a \in \mathbb{R}^{W}$ that maximizes the expected number of users that click on the advertisement at least once:

\begin{equation}
\begin{aligned}
    \max_{a} \quad & \sum_{u=1}^{U}(1 - \prod_{w=1}^{W} (1 - a_w \cdot \hat{y}_{wu})) \notag \\
    \text{s.t.} \quad & a\in \{0, 1\}^W
\end{aligned}
\end{equation}

\subsection{Details in Portfolio Optimization Problem}

For the portfolio optimization problem, we use historical daily stock return data from 2004 to 2017 for $N=50$ stocks \cite{shah2022decision}.
Using this historical daily stock return data, we predict the future stock price $y \in \mathbb{R}^{N}$.
Subsequently, we solve for the portfolio weight $a \in \mathbb{R}^{N}$, aiming to maximize the expected return given the covariance matrix $\Sigma$ and the risk aversion coefficient $\gamma$:

\begin{equation}
\begin{aligned}
    \max_{a} \quad & y^\top a - \gamma \cdot a^\top \Sigma a \notag \\
    \text{s.t. } \quad & \sum_{i=1}^{N} a_i = 1
\end{aligned}
\end{equation}
where we set $\gamma=0.1$.

\section{Additional Results}

\subsection{Fake Target Variants Results for Budget Allocation Problem}
\label{subsec:appen-diff-faketargets}
Here, we show the results of budget allocation with varying numbers of fake targets.
We tested problems having $\{0,5,50,500\}$ fake targets, where the problem becomes more challenging as the fake targets increase.
We use 16 samples for surrogate loss models.
The results are presented in Table \ref{tab:fake-target}.
The normalized test regret is lower the better, and we bold-lettered the results for each fake target experiment.

\begin{table*}[ht]

\renewcommand{\arraystretch}{1.3}
    \begin{center}
    \begin{small}
    \begin{sc}
        \begin{tabular}{lcccc}
        \toprule
        \multirow{2}{*}{\textbf{Method}}       & \multicolumn{4}{c}{\# Fake Targets} \\
        \cline{2-5}
                 & 0 & 5 & 50 & 500 \\
        \midrule
        \texttt{PFL}      & 0.128 $\pm$ 0.018 & 0.193 $\pm$ 0.028 & \textbf{0.242 $\pm$ 0.026} & 0.513 $\pm$ 0.016 \\
        \texttt{DFL}      & \textbf{0.064 $\pm$ 0.016} & 0.175 $\pm$ 0.026 & 0.377 $\pm$ 0.012 & 0.502 $\pm$ 0.011 \\
        \texttt{LODL-DQ}  & 0.133 $\pm$ 0.020 & 0.184 $\pm$ 0.017 & 0.326 $\pm$ 0.018 & 0.509 $\pm$ 0.008 \\
        \texttt{LANCER}   & 0.093 $\pm$ 0.013 & 0.185 $\pm$ 0.026 & 0.282 $\pm$ 0.031 & 0.490 $\pm$ 0.010 \\
        \texttt{EGL-WMSE} & 0.189 $\pm$ 0.035 & 0.210 $\pm$ 0.033 & 0.340 $\pm$ 0.027 & 0.518 $\pm$ 0.007 \\
        \texttt{EGL-DQ}   & 0.068 $\pm$ 0.021 & 0.155 $\pm$ 0.037 & 0.289 $\pm$ 0.016 & 0.496 $\pm$ 0.016 \\
        \texttt{LCGLN}     & 0.090 $\pm$ 0.013 & \textbf{0.112 $\pm$ 0.024} & 0.272 $\pm$ 0.015 & \textbf{0.478 $\pm$ 0.007} \\
        \bottomrule
        \end{tabular}
    \end{sc}
    \end{small}
    \end{center}
    \caption{
    The table presents normalized test regret $\mathcal{R}_{test} / \mathcal{R}_{worst}$ with standard error mean (SEM) tested on the budget allocation problem.
    The metric is \emph{lower the better} with 0 meaning optimal.
    We tested for varying fake target numbers of $\{0,5,50,500\}$.
    Surrogate loss models were trained with 16 samples.
    The best-performing results for each number of fake targets are bold-lettered.
    We evaluate our LCGLN method against all baselines: PFL, DFL, and both local and global surrogate loss models.
    }
    \label{tab:fake-target}
\end{table*}

\subsection{Sample Size Variants Results}
\label{subsec:appen-diff-smpsize}
We present our experimental results using different sample sizes of $\{2,4,8,16,32\}$ for each problem and methodology.
The results of normalized test regret for inventory stock, budget allocation, and portfolio optimization problem are in Table \ref{tab:sample-size-variants-intentory}, \ref{tab:sample-size-variants-budget}, and \ref{tab:sample-size-variants-portfolio} respectively.
Note that the results of PFL and DFL remain unchanged regardless of the number of samples since they do not sample predictions.
We bold-lettered the best-performing results for each sample size.
 

\begin{table*}[ht]

\renewcommand{\arraystretch}{1.3}
    \begin{center}
    \begin{small}
    \begin{sc}
        \begin{tabular}{lccccc}
        \toprule
        \multirow{2}{*}{\textbf{Method}}   & \multicolumn{5}{c}{\# Samples} \\
        \cline{2-6}
            & 2     & 4      & 8     & 16    & 32   \\
        \midrule
           \texttt{PFL}      & 0.242 $\pm$ 0.005 & 0.242 $\pm$ 0.005 & 0.242 $\pm$ 0.005 & 0.242 $\pm$ 0.005 & 0.242 $\pm$ 0.005 \\
           \texttt{DFL}      & 0.228 $\pm$ 0.002 & 0.228 $\pm$ 0.002 & 0.228 $\pm$ 0.002 & 0.228 $\pm$ 0.002 & 0.228 $\pm$ 0.002 \\
           \texttt{LODL-DQ}  & 0.371 $\pm$ 0.005 & 0.365 $\pm$ 0.003 & 0.373 $\pm$ 0.006 & 0.375 $\pm$ 0.005 & 0.378 $\pm$ 0.007 \\
           \texttt{LANCER}   & \textbf{0.182 $\pm$ 0.004} & \textbf{0.182 $\pm$ 0.004} & \textbf{0.182 $\pm$ 0.004} & \textbf{0.182 $\pm$ 0.004} & 0.182 $\pm$ 0.004 \\
           \texttt{EGL-WMSE} & 0.383 $\pm$ 0.007 & 0.372 $\pm$ 0.004 & 0.371 $\pm$ 0.006 & 0.365 $\pm$ 0.003 & 0.371 $\pm$ 0.002 \\
           \texttt{EGL-DQ}   & 0.376 $\pm$ 0.005 & 0.368 $\pm$ 0.003 & 0.367 $\pm$ 0.002 & 0.366 $\pm$ 0.001 & 0.369 $\pm$ 0.007 \\
           \texttt{LCGLN}     & 0.237 $\pm$ 0.005 & 0.233 $\pm$ 0.004 & 0.224 $\pm$ 0.004 & 0.208 $\pm$ 0.007 & \textbf{0.174 $\pm$ 0.002} \\
        \bottomrule
        \end{tabular}
    \end{sc}
    \end{small}
    \end{center}
    \caption{
    The table presents normalized test regret $\mathcal{R}_{test} / \mathcal{R}_{worst}$ with standard error mean (SEM) tested on the inventory stock problem.
    The metric is \emph{lower the better} with 0 meaning optimal.
    We tested for varying sample sizes of $\{2,4,8,16,32\}$.
    The best-performing results for each sample size are bold-lettered.
    We evaluate our LCGLN method against all baselines: PFL, DFL, and both local and global surrogate loss models.
    LCGLN outperformed the baselines as the number of samples reached 32.
    }
    \label{tab:sample-size-variants-intentory}
\end{table*}
\begin{table*}[ht]
\renewcommand{\arraystretch}{1.3}
    \begin{center}
    \begin{small}
    \begin{sc}
        \begin{tabular}{lccccc}
        \toprule
        \multirow{2}{*}{\textbf{Method}}   & \multicolumn{5}{c}{\# Samples} \\
        \cline{2-6}
            & 2     & 4      & 8     & 16    & 32   \\
        \midrule
           \texttt{PFL}      & 0.513 $\pm$ 0.016 & 0.513 $\pm$ 0.016 & 0.513 $\pm$ 0.016 & 0.513 $\pm$ 0.016 & 0.513 $\pm$ 0.016 \\
           \texttt{DFL}      & 0.502 $\pm$ 0.011 & 0.502 $\pm$ 0.011 & 0.502 $\pm$ 0.011 & 0.502 $\pm$ 0.011 & 0.502 $\pm$ 0.011 \\
           \texttt{LODL-DQ}  & 0.533 $\pm$ 0.008 & 0.523 $\pm$ 0.011 & 0.509 $\pm$ 0.011 & 0.509 $\pm$ 0.008 & 0.503 $\pm$ 0.020 \\
           \texttt{LANCER}   & \textbf{0.490 $\pm$ 0.010} & \textbf{0.490 $\pm$ 0.010} & \textbf{0.490 $\pm$ 0.010} & 0.490 $\pm$ 0.010 & 0.490 $\pm$ 0.010 \\
           \texttt{EGL-WMSE} & 0.537 $\pm$ 0.008 & 0.529 $\pm$ 0.007 & 0.520 $\pm$ 0.006 & 0.518 $\pm$ 0.007 & 0.510 $\pm$ 0.013 \\
           \texttt{EGL-DQ}   & 0.520 $\pm$ 0.013 & 0.514 $\pm$ 0.015 & 0.499 $\pm$ 0.012 & 0.496 $\pm$ 0.016 & 0.492 $\pm$ 0.005 \\
           \texttt{LCGLN}     & 0.529 $\pm$ 0.016 & 0.496 $\pm$ 0.020 & 0.492 $\pm$ 0.008 & \textbf{0.478 $\pm$ 0.007} & \textbf{0.468 $\pm$ 0.009} \\
        \bottomrule
        \end{tabular}
    \end{sc}
    \end{small}
    \end{center}
    \caption{
    The table presents normalized test regret $\mathcal{R}_{test} / \mathcal{R}_{worst}$ with standard error mean (SEM) tested on the budget allocation problem with 500 fake targets.
    The metric is \emph{lower the better} with 0 meaning optimal.
    We tested for varying sample sizes of $\{2,4,8,16,32\}$.
    The best-performing results for each sample size are bold-lettered.
    We evaluate our LCGLN method against all baselines: PFL, DFL, and both local and global surrogate loss models.
    LCGLN outperformed the baselines as the number of samples reached 16.
    }
    \label{tab:sample-size-variants-budget}
\end{table*}
\begin{table*}[ht]
\renewcommand{\arraystretch}{1.3}
    \begin{center}
    \begin{small}
    \begin{sc}
        \begin{tabular}{lccccc}
        \toprule
        \multirow{2}{*}{\textbf{Method}}   & \multicolumn{5}{c}{\# Samples} \\
        \cline{2-6}
            & 2     & 4      & 8     & 16    & 32   \\
        \midrule
           \texttt{PFL}      &  0.189 $\pm$ 0.002 & 0.189 $\pm$ 0.002 &          0.189 $\pm$ 0.002 & 0.189 $\pm$ 0.002 & 0.189 $\pm$ 0.002 \\
           \texttt{DFL}      & 0.187 $\pm$ 0.002 & 0.187 $\pm$ 0.002 & 0.187 $\pm$ 0.002 & 0.187 $\pm$ 0.002 & 0.187 $\pm$ 0.002 \\
           \texttt{LODL-DQ}  &  0.214 $\pm$ 0.003 & 0.212 $\pm$ 0.004 & 0.202 $\pm$ 0.002 & 0.195 $\pm$ 0.002 & 0.193 $\pm$ 0.002 \\
           \texttt{LANCER}   & 0.228 $\pm$ 0.006 & 0.228 $\pm$ 0.006 & 0.228 $\pm$ 0.006 & 0.228 $\pm$ 0.006 & 0.228 $\pm$ 0.006 \\
           \texttt{EGL-WMSE} & 0.191 $\pm$ 0.002 & 0.190 $\pm$ 0.003 & 0.189 $\pm$ 0.002 & 0.188 $\pm$ 0.003 & 0.187 $\pm$ 0.001 \\
           \texttt{EGL-DQ}   &  0.260 $\pm$ 0.005 & 0.257 $\pm$ 0.004 & 0.257 $\pm$ 0.004 & 0.256 $\pm$ 0.003 & 0.256  $\pm$ 0.002 \\
           \texttt{LCGLN}     & \textbf{0.186 $\pm$ 0.000} & \textbf{0.185 $\pm$ 0.000} & \textbf{0.185 $\pm$ 0.000} & \textbf{0.185 $\pm$ 0.000} & \textbf{0.185 $\pm$ 0.000} \\
        
        \bottomrule
        \end{tabular}
    \end{sc}
    \end{small}
    \end{center}
    \caption{
    The table presents normalized test regret $\mathcal{R}_{test} / \mathcal{R}_{worst}$ with standard error mean (SEM) tested on the portfolio optimization problem.
    The metric is \emph{lower the better} with 0 meaning optimal.
    We tested for varying sample sizes of $\{2,4,8,16,32\}$.
    The best-performing results for each sample size are bold-lettered.
    We evaluate our LCGLN method against all baselines: PFL, DFL, and both local and global surrogate loss models.
    LCGLN outperformed the baselines for all sample sizes.
    }
    \label{tab:sample-size-variants-portfolio}
\end{table*}

\end{document}